\documentclass[11pt,draftcls,onecolumn]{IEEEtran}
\usepackage{amssymb,amsfonts,amsthm, lipsum}

\usepackage{times}

\usepackage{subfigure}

\usepackage{multirow}

\usepackage[hyperindex,breaklinks]{hyperref}

\usepackage{cite}

\ifCLASSINFOpdf
   \usepackage[pdftex]{graphicx}
   \graphicspath{{pdf/}{jpeg/}}
   \DeclareGraphicsExtensions{.pdf,.jpeg,.png}
\else
   \usepackage[dvips]{graphicx}
   \DeclareGraphicsExtensions{.eps}
\fi
\usepackage[cmex10]{amsmath}
\usepackage{algorithm}
\usepackage{algorithmic}

\usepackage{array}
\usepackage{url}

\theoremstyle{plain} \newtheorem{thm}{Theorem}
\newtheorem{cor}[thm]{Corollary} \newtheorem{lem}[thm]{Lemma}

\theoremstyle{definition} 

\theoremstyle{remark} 

\newtheorem{lemma}{Lemma}

  \newenvironment{Proof}{\noindent{\bf Proof} \ }{\QED}\smallskip
\newcommand\QED{\newline \rightline{$\blacksquare$} \bigskip}

  \newenvironment{Proof thm4}{\noindent{\bf Proof of Theorem 4} \ }{\QED}\smallskip

  \newenvironment{Proof thm5}{\noindent{\bf Proof of Theorem 5} \ }{\QED}\smallskip

  \newenvironment{Proof thm6}{\noindent{\bf Proof of Theorem 6} \ }{\QED}\smallskip

\newcommand\Rt{\mathcal{R}_{L,T,\lambda}}
\newcommand\Rp{\mathcal{R}_{L,P,\lambda}}
\newcommand\RT{\mathcal{R}_{L,T}}
\newcommand\RP{\mathcal{R}_{L,P}}

\newcommand\h{\mathcal{H}}
\newcommand\e{\epsilon_n}
\newcommand\X{\mathcal{X}}
\newcommand\x{\mathbf{X}}
\newcommand\F{\mathcal{F}}

\usepackage{setspace}

\begin{document}
%
\title{A Rank-SVM Approach to Anomaly Detection}
%
%
%

\author{Jing~Qian,~\IEEEmembership{Student~Member,~IEEE,}
Jonathan~Root,~\IEEEmembership{Student~Member,~IEEE,}
Venkatesh~Saligrama,~\IEEEmembership{Senior~Member,~IEEE}
        and~Yuting~Chen,~\IEEEmembership{Student~Member,~IEEE,}
\thanks{J. Qian and Y. Chen are with the Division of Systems Engineering, Boston University, 15 Saint Mary's St., Brookline, MA 02446 USA. E-mail: \{jingq,yutingch\}@bu.edu.}
\thanks{V. Saligrama is with the Department of Electrical and Computer Engineering, Boston University, 8 Saint Mary's St., Boston,
MA, 02215 USA. E-mail: srv@bu.edu.}
\thanks{J. Root is with the Department of Mathematics and Statistics, Boston University, 8 Saint Mary's St., Boston,
MA, 02215 USA. E-mail: jroot@bu.edu.}
}

%
%

\markboth{A New One-class SVM for Anomaly Detection}{A New One-class SVM for Anomaly Detection}
%


\maketitle

\begin{abstract}
We propose a novel non-parametric adaptive anomaly detection algorithm for high dimensional data based on rank-SVM. Data points are first ranked based on scores derived from nearest neighbor graphs on $n$-point nominal data. We then train a rank-SVM using this ranked data. A test-point is declared as an anomaly at $\alpha$-false alarm level if the predicted score is in the $\alpha$-percentile. The resulting anomaly detector is shown to be asymptotically optimal and adaptive in that for any false alarm rate $\alpha$, its decision region converges to the $\alpha$-percentile level set of the unknown underlying density. In addition we illustrate through a number of synthetic and real-data experiments both the statistical performance and computational efficiency of our anomaly detector.
\end{abstract}


%
\IEEEpeerreviewmaketitle

\section{Introduction}\label{sec:intro}
Anomaly detection is the problem of identifying statistically significant deviations of data from expected normal behavior. It has found wide applications
in many areas such as credit card fraud detection, intrusion detection for cyber security, sensor networks and video surveillance \cite{ref:anomaly_detection_survey}\cite{ref:AD_survey_hodge}.
In this paper we focus on the setting of point-wise anomaly detection. At training time only a set of nominal examples drawn i.i.d. from an unknown ``nominal'' distribution are given. Note that nominal density or distribution refer to the distribution of nominal samples in this paper, not the Gaussian distribution.
Our objective is to learn an anomaly detector that maximizes the detection power subject to some false-alarm rate constraint.

Existing approaches of point-wise anomaly detection can be divided into two categories, namely parametric and non-parametric methods.
Classical parametric methods \cite{ref:para_1993} for anomaly detection assume a family of functions that characterize the nominal density. Parameters are then estimated from training data by minimizing some loss function. While these methods provide a statistically justifiable solution when the assumptions hold true, they are likely to suffer from model mismatch and lead to poor performance.

Popular non-parametric approaches include one-class support vector machines (SVM) \cite{ref:oc_svm2001}, and various density-based algorithms \cite{ref:MV_2006, ref:GEM_2006, ref:Manqi2009, ref:Jing2012, ref:knn_2011}.
The kernel one-class SVM algorithm attempts to find a decision boundary by mapping the nominal data into a high-dimensional kernel space, separating the image from the origin with maximum margin.
While one-class SVM is computationally efficient it does not directly incorporate density information, and can exhibit poor control over false alarm rates.
Density-based methods such as minimum volume set estimation \cite{ref:MV_2006,ref:MV_2010} and geometric entropy minimization (GEM) \cite{ref:GEM_2006} involve explicitly approximating high-dimensional quantities such as the multivariate density function or the minimum volume set boundaries. This can be computationally prohibitive for high dimensional problems.
Nearest neighbor-based approaches \cite{ref:Manqi2009, ref:knn_2011, ref:Jing2012} propose to estimate the $p$-value function through some statistic based on $k$ nearest neighbor ($k$-NN) distances within the graph constructed from nominal points. While allowing flexible control over false alarm rates and often providing better performance than one-class SVM, these approaches usually require expensive computations at test stage, such as calculating the $k$-NN distance of the test point, which makes them inapplicable for tasks requiring real-time processing.

In this paper we propose a novel RankSVM based anomaly detection algorithm that combines the computational efficiency of the simple one-class SVM approach with the statistical performance of nearest neighbor-based methods. Our approach learns a ranker over nominal samples through a ``supervised'' learning-to-rank step, for which we adopt the pair-wise RankSVM algorithm. Nominal density information based on $k$-NN distances is directly incorporated as input pairs in this pair-wise learning-to-rank step. For each input pair $(x_i,x_j)$ a binary value is assigned with zero denoting the fact that point $x_i$ is more of an outlier relative to point $x_j$ and one representing the opposite scenario. The learning step trains a ranker which predicts how anomalous a point is.
At the test stage our detector labels a test point as an anomaly if the predicted rank score is in the $\alpha$-percentile among the training data. We then show asymptotic consistency and present a finite-sample generalization result of our ranking-based anomaly detection framework. We conduct experiments on both synthetic and real data sets to demonstrate superior performance in comparison to other methods.

We summarize the advantages of our proposed anomaly detection approach below.

\begin{itemize}
\item Computational Efficiency: During test stage, our approach only needs to evaluate an SVM-type function on the test point, similar to the simple one-class SVM approach. In contrast nearest neighbor-based methods generally require distances between the test point and training points to determine whether or not the test point is anomalous.
\item Statistical Performance: Our discriminative learning-to-rank step leads to a ranking-based anomaly detection framework that is asymptotically consistent. We can also guarantee a finite sample bound on the empirical false alarm rate of our decision rule.
\item Adaptivity: Our method adapts to any false alarm level because it can asymptotically approximate different level-sets of the underlying density function. While the threshold parameter can be modified for one-class SVM to obtain detectors for different false alarm levels this does not often result in optimal performance.
\end{itemize}

%
%

The rest of the paper is organized as follows. In Section 2 we introduce the problem setting and the motivation. Detailed algorithms are described in Section 3. The asymptotic and finite-sample analyses are provided in Section 4. Synthetic and real experiments are reported in Section 5. Section 6 concludes the paper.

\section{Problem Setting and Motivation}
We closely follow the setting of \cite{ref:oc_svm2001}.
Let $\mathbf{x}=\lbrace x_1, x_2, ..., x_n\rbrace $, $x_i\in X=\mathbb{R}^d$, be a given set of nominal points sampled i.i.d from an unknown density $f$ with compact support in $\mathbb{R}^d$. Let $P$ be the corresponding probability measure.
We are interested in estimating the $(1-\alpha)$-percentile minimum volume set with reference measure $\mu$:
\begin{equation}\label{eq:mv_set}
  U_{1-\alpha} = \arg \min_{U} \left\{ \mu(U): \,\, P(U) \geq 1-\alpha, \, U \, \text{measurable} \right\}.
\end{equation}
The most common choice of $\mu$ is Lebesgue measure \cite{ref:oc_svm2001}, in which case $U_{1-\alpha}$ represents the minimum volume set that captures at least a fraction $1-\alpha$ of the probability mass.
Its meaning in outlier/anomaly detection is that for a given test point $\eta$, the following decision rule naturally maximizes the detection power
\footnote{ Such a rule maximizes the detection rate with respect to the reference measure $\mu$. Usually no anomalous samples are available during training; the common choice of Lebesgue measure as $\mu$ corresponds to assuming uniform anomaly distribution. It is shown that the decision rule \ref{eq:decision_mvset} is the uniformly most powerful (Neyman-Pearson) test \cite{ref:Manqi2009}. }
while controlling the false alarm rate at $\alpha$, $P(\text{declare} \, H_1|H_0) \leq \alpha$,
\begin{equation}\label{eq:decision_mvset}
D(\eta) =
  \left\{
   \begin{array}{cc}
   \text{nominal} & \,\, \eta \in U_{1-\alpha}  \\
   \text{anomaly} & \text{otherwise} \\
   \end{array}
  \right.
\end{equation}

The following lemma restates and simplifies the above optimal decision rule in terms of the $p$-value function, which will be the main focus of this paper.
\begin{lem}\label{lem_p_mvset}
Assume the density $f$ has no ``flat region'' on the support, $\mu\left( \{ x:f(x)=\gamma \} \right)=0,\forall \gamma>0$. Then the decision rule Eq.(\ref{eq:decision_mvset}) is equivalent to:
\begin{equation}\label{eq:decision_pvalue}
D(\eta) =
  \left\{
   \begin{array}{cc}
   \text{anomaly} & \, p(\eta) \leq \alpha  \\
   \text{nominal} & \, \text{otherwise} \\
   \end{array}
  \right.
\end{equation}
where $p(\eta)$ is the $p$-value function defined as:
\begin{equation}
p(\eta) = P
\left(
x:f(x) \leq f(\eta)
\right)
 = \int_{\lbrace x: f(x) \leq f(\eta)\rbrace } f(x)\, dx
\end{equation}
\end{lem}

The proof is straight forward. \cite{ref:MV_2006,ref:levelset_mvset_2003} has shown that given the non-flat assumption, the minimum volume set coincides with the $(1-\alpha)$-quantile density level set:
\begin{equation}\label{eq:mvset_levelset}
  U_{1-\alpha} = \left\{ x: \, f(x)\geq \gamma_{1-\alpha} \right\} = \left\{ x: \, p(x)\geq 1-\alpha \right\}.
\end{equation}
$\eta \in U_{1-\alpha}$ thus yields $p(\eta) \geq 1-\alpha$, which leads to Eq.(\ref{eq:decision_pvalue}).

The main concern now is to estimate $p(\eta)$. It is worth mentioning that instead of estimating one particular minimum volume set as in oc-svm \cite{ref:oc_svm2001}, we aim to estimate $p$-value function which yields different decision regions with different values of $\alpha$. This point will be illustrated later.
We are motivated to learn a ranker on nominal samples based on the following lemma:

\begin{lem}\label{lem_order_consistency}
Assume $G$ is any function that gives the same order relationship as the density: $G(x_i)>G(x_j)$, $\forall x_i\neq x_j$ such that $f(x_i)>f(x_j)$. Then as $n\rightarrow\infty$, the rank of $\eta$, $r(\eta)$, converges to $p(\eta)$:
\begin{equation}\label{eq:rank}
    r(\eta) = \frac{1}{n}\sum_{j=1}^{n} \textbf{1}_{\{ G(x_j)\leq G(\eta) \}} \rightarrow p(\eta).
\end{equation}
\end{lem}
By assumption we have $\textbf{1}_{\{ G(x)\leq G(y) \}} = \textbf{1}_{\{ f(x)\leq f(y) \}}$; the proof is just an application of the law of large numbers.

The lemma shows that if some statistic $G$ preserves the ordering of the density $f$ on the data set, the rank $r$ asymptotically converges to the $p$-value function $p$.
There are a number of choices for the statistic $G$. One could of course compute the density estimates as $G$ and plug into Eq.(\ref{eq:rank}).
On the other hand, nearest neighbor-based approaches \cite{ref:Manqi2009, ref:knn_2011, ref:Jing2012} propose simple statistics based on the $k$-nearest neighbor ($k$-NN) distance or the $\epsilon$-neighborhood size among the nominal set $\{x_1,\cdots,x_n\}$ as surrogates, which are shown to be asymptotically consistent as in Eq.(\ref{eq:rank}).
Other techniques, such as $k$-NN geodesic distance of Isomap \cite{ref:isomap_2000} or $k$-NN distance after Local Linear Embedding (LLE) \cite{ref:LLE_2000}, can adapt to intrinsic dimension instead of ambient dimension. Recently $k$-NN approaches that is customized to each data point has also been shown to achieve this \cite{ref:adapt_2013}.
It should be noted that although choosing $G$ is important, this is not our main focus. In fact our method would work in conjunction with these techniques. For simplicity of exposition we restrict ourselves to $k$-NN statistic as $G$ in this paper.

The main issue is that we would like to avoid computing the $k$-NN distance statistic (or other complicated $G$) for $\eta$ during test stage, because the complexity would grow as $\Omega(dn+n\log n)$, which can be prohibitive for real-time applications.
To this end, we are compelled to learn a simple scoring function $g:X\to \mathbb{R}$, that best respects the observed pair-wise preference relationship given by $G$,
$\{(i,j) : r(x_i) > r(x_j)\}$.
This is achieved via a supervised pair-wise learning-to-rank framework. The inputs are preference pairs encoding the ground-truth order relationship information.
In our setting, we generate preference pairs based on the average $k$-NN distance statistic, which has been shown \cite{ref:Jing2012} to be robust and to have good convergence rates. In this way, nominal density information is incorporated in the input pairs.

Now that we have the preference pairs, the next step is to learn a ranker based on minimizing the pair-wise disagreement loss function.
In our work we adopt the rank-SVM method to obtain a ranker $g:X\to \mathbb{R}$.
Intuitively, $g(\cdot)$ scores how ``nominal'' the point $x$ is, and is simple to evaluate relative to density-based counterparts. This leads to reduced complexity during the test stage. $g$ is then adopted in place of $G$ to compute the rank $r(\eta)$ according to Eq.(\ref{eq:rank}), which only requires a bisection search among sorted $\{r(x_i),\, i=1,\ldots,n\}$ of nominal samples.

\section{Rank-Based Anomaly Detection Algorithm}\label{sec:main_algo}

\subsection{Anomaly Detection Algorithm}

We describe the main steps of the algorithm:

\subsubsection{Rank Computation}

For each nominal training sample $x_i$, let $D_{(i)}(x)$ be the distance to its $i$th nearest neighbor. Then,
\begin{equation}\label{eq:aKnn}
G(x) = - \frac{1}{K}\sum^{K}_{i=1}D_{(i)}(x)
\end{equation}
for some suitable $K$.
We plug $G$ into Eq.(\ref{eq:rank}) and compute ranks $r(x_i)$ of nominal points.

\subsubsection{Learning-to-Rank}

From Step 1 our training set is now $\lbrace(x_i, r(x_i))\rbrace$. Next we want to learn a ranker $g(\cdot)$ so that it outputs an ordinal value $g(x_i)$ for each $x_i$ which maximally preserves the ordering of $r(x_i)$.
We adopt the pairwise learning-to-rank framework, where the input is a collection of preference pairs, $\mathcal{P}$, where each input pair $(i,j)\in \mathcal{P}$ represents $r(x_i)\geq r(x_j)$. The goal is to minimize some loss function, for example, the weighted pairwise disagreement loss (WPDL),
\begin{equation}\label{eq:WPDL}
  l(r,g) = \sum_{(i,j)\in \mathcal{P}} w_{ij}\textbf{1}_{\{g(x_i)<g(x_j)\}}
\end{equation}
We adopt the rank-SVM algorithm to train our ranker $g$ with equal weight $w_{ij}=1$ for all pairs $(i,j)\in\mathcal{P}$, and solve the following optimization problem:
\begin{eqnarray}\label{eq:ranksvm_standard}
  \min_{\omega,\xi_{ij}}: \, && \,\, \frac{1}{2} ||\omega||^2 + C \sum_{(i,j)\in\mathcal{P}} \xi_{ij} \\
\nonumber
  s.t. \, &&  \,\, \langle \omega,\, \Phi(x_i)-\Phi(x_j) \rangle \geq 1 - \xi_{ij}, \,\,\,\, \forall (i,j)\in \mathcal{P}  \\
\nonumber
          &&  \,\,  \xi_{ij} \geq 0
\end{eqnarray}
where $\Phi:X\to H$ is a mapping into a reproducing kernel Hilber space $H$ with inner product $\langle \cdot,\cdot \rangle$. Rank-SVM minimizes WPDL with indicator replaced by hinge loss. Details about rank SVM can be found in \cite{ref:ranksvm}.

\subsubsection*{Remark 1}
Given the ranks $r(x_i)$ of $n$ nominal samples, practically we find that generating all $n \choose 2$ preference pairs for the rank-SVM algorithm often leads to poor detection performance due to overfitting, not to mention the high training complexity. In our experiments the following scheme is adopted: we first quantize all the ranks to $m$ levels $r_q(\cdot)\in \lbrace 1,..,m \rbrace$. A preference pair $(i, j)\in\mathcal{P}$ is generated for every $r_q(x_i) > r_q(x_j)$, indicating that samples near $x_j$ are ``less nominal'', thus ``more anomalous'' than those near $x_i$. This scheme with a relatively small $m$ significantly reduces the number of input pairs to the rank-SVM algorithm and the training time. It results in better empirical performance as well. While this raises the question of choosing $m$ we find $m=3$ works fairly well in practice and we fix this in all of our experiments in Sec.\ref{sec:exp}.
Other similar schemes can be used to select preference pairs, for example, only pairs $(i,j)$ with significant rank differences $r(x_i)-r(x_j)\geq \tau$ are input to ranking-SVM.

\subsubsection*{Remark 2} We adopt the RBF kernel $K(x_i,x_j) = \exp\left( -\dfrac{(x_i-x_j)^2}{\sigma^2} \right)$ for rank-SVM. The algorithm parameter $C$ and RBF kernel bandwidth $\sigma$ can be selected through cross validation, since this rank-SVM step is a supervised learning procedure based on input pairs.

\subsubsection{Prediction}

At test time, the ordinal value for $\eta$, $g(\eta)$ is first computed.
Then the rank $r(\eta)$ is estimated using Eq.(\ref{eq:rank}) by replacing $G(\cdot)$ with $g(\cdot)$. If $r(\eta)$ falls under the false alarm level $\alpha$, anomaly is declared.

Our algorithm is summarized as follows:

\noindent\rule[0.5ex]{\linewidth}{1pt}
\noindent\textbf{Algorithm 1: Ranking Based Anomaly Detection (rankAD)}

\noindent\rule[0.5ex]{\linewidth}{1pt}
\noindent\textbf{1. Input:}

\noindent nominal training data $\mathbf{x}={\lbrace x_1, x_2, ..., x_n \rbrace}$, desired false alarm level $\alpha$, and test point $\eta$

\noindent\textbf{2. Training Stage:}

\noindent(a) Calculate ranks $G(x_i)$ and thus $r(x_i)$ for each nominal sample $x_i$, using Eq.(\ref{eq:aKnn}) and Eq.(\ref{eq:rank}). \\
\noindent(b) Quantize the ranks $r(x_i)$ into $3$ levels: $r_q(x_i) \in \lbrace 1,2,3 \rbrace$. Generate  preference pairs $(x_i, x_j)$ whenever their quantized levels satisfy $r_q(x_i) > r_q(x_j)$. \\
\noindent(c) Train a ranker $g(\cdot)$ through RankSVM. \\
\noindent(d) Compute $g(x_i)$ of $x_1,\ldots,x_n$, and sort these values.

\noindent\textbf{3. Testing Stage:}

\noindent(a) Evaluate $g(\eta)$ for test point $\eta$.\\
\noindent(b) Compute the rank $r(\eta)$ according to Eq.(\ref{eq:rank}), replacing $G(\cdot)$ with $g(\cdot)$.\\
\noindent(c) Declare $\eta$ as anomalous if $r(\eta) \leq \alpha$.

\noindent\rule[0.5ex]{\linewidth}{1pt}

\subsection{Comparison With State-of-the-art Algorithms}
We provide comparison of our approach against one-class SVM and density-based algorithms in terms of false alarm control and test stage complexity.


\subsubsection{False alarm control}
One-class SVM does not have any natural control over the false alarm rate. Usually the parameter $\nu$ is varied for a different false alarm level, requiring re-solving the optimization problem. This is because one-class SVM aims at approximating one level set at a time.
While our method also involves SVM learning step, our approach is substantially different from one-class SVM. Our ranker $g$ from the rank-SVM step simultaneously approximates multiple level sets. The normalized score Eq.(\ref{eq:rank}), takes values in $[0,1]$, and converges to the $p$-value function. Therefore we get a handle on the false alarm rate. So null hypothesis can be rejected at different levels simply by thresholding $r(\eta)$.

\begin{figure*}[t!]
\begin{centering}
\begin{minipage}[t]{.44\textwidth}
\includegraphics[width = 1\textwidth]{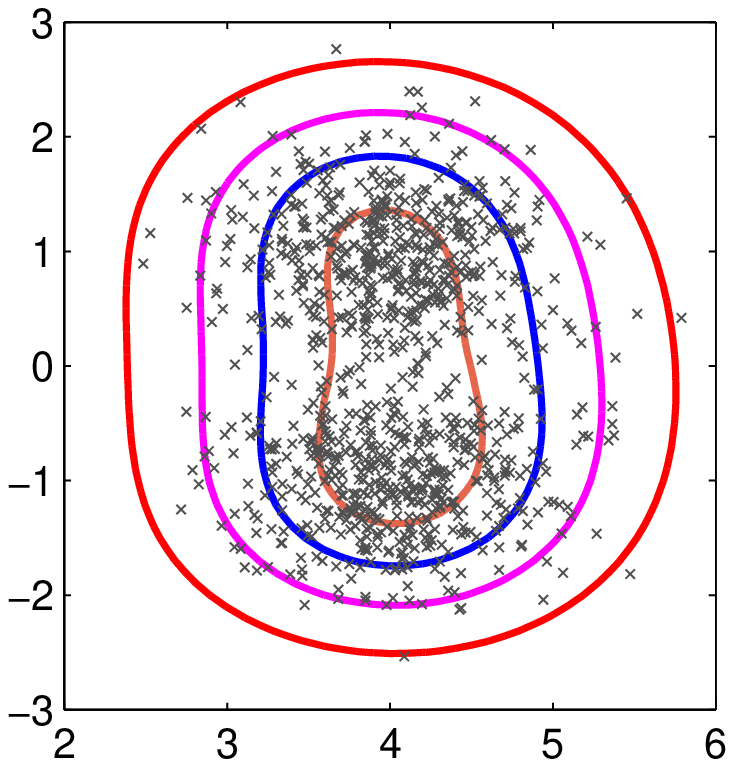}
\makebox[7 cm]{(a) Level curves of oc-svm}\medskip
\end{minipage}
\begin{minipage}[t]{.44\textwidth}
\includegraphics[width = 1\textwidth]{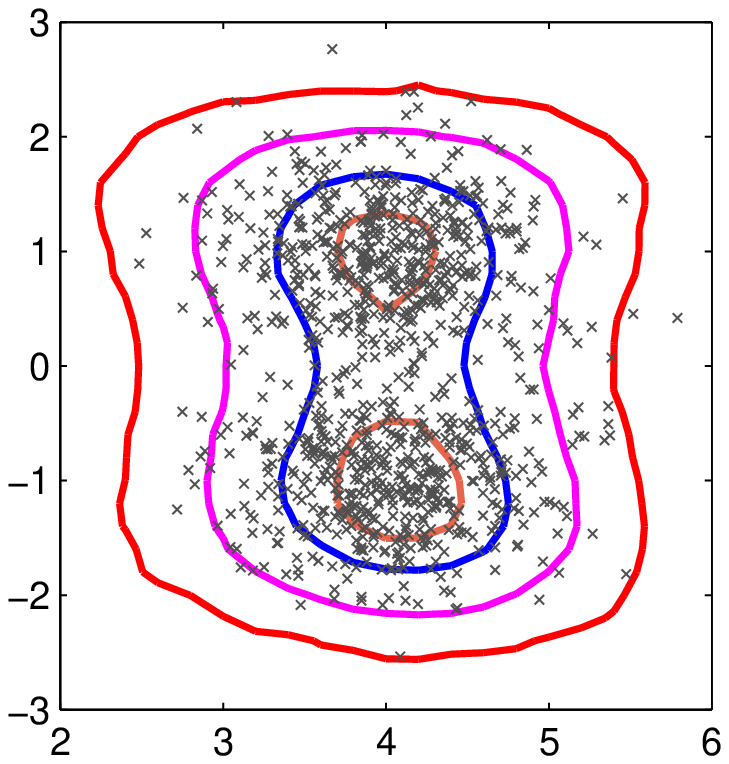}
\makebox[7 cm]{(b) Level curves of rank-SVM}\medskip
\end{minipage}
\caption{ Level curves of one-class SVM and rank-SVM. 1000 i.i.d. samples are drawn from a 2-component Gaussian mixture density. Let $g_{oc},g_{R}$ are one-class SVM ($\nu=0.03,\sigma=1.5$) and our rankSVM ($m=3,C=1,\sigma=1.5$) predictors respectively. (a) shows level curves obtained by varying the offset $c_{oc}$ for $g_{oc}(x)=c_{oc}$. Only the outmost curve ($c_{oc}=0$) approximates the oracle density level set well while  the inner curves ($c_{oc}>0$) appeared to scaled version of outermost curve. (b) shows level curves obtained by varying $c_R$ for $g_{R}(x)=c_{R}$. Notice that the inner most curve approximates peaks of the mixture density. }
\end{centering}
\label{fig:level_curve}
\end{figure*}

{\bf Toy Example:} We present a simple example in Fig.1 to demonstrate this point. The nominal density $ f \sim 0.5 \mathcal{N} \left(
\left[ 4; 1 \right],
0.5 I
 \right)
+
0.5 \mathcal{N} \left(
\left[ 4 ; -1 \right],
0.5 I
 \right)
$.
One-class SVM using RBF kernels ($\sigma=1.5$) is trained with parameter $\nu=0.03$, to yield a decision function $g_{oc}(\cdot)$. The standard way is to claim anomaly when $g_{oc}(x)<0$, corresponding to the outmost orange curve in (a). We then plot different level curves by varying $c_{oc}>0$ for $g_{oc}(x)=c_{oc}$, which appear to be scaled versions  of the orange curve. Intuitively, this is because one-class SVM with parameter $\nu$ aims to separate approximately $1-\nu$ fraction of nominal points from the origin in RKHS with maximum-margin principle, and only focuses on points near the boundary. Asymptotically it is known to approximate one density level set well \cite{ref:oc_svm2001}. For a different $\alpha$, one-class SVM needs re-training with a different $\nu$. On the other hand, we also train rank-SVM with $m=3,C=1,\sigma=1.5$ and obtain the ranker $g_R(\cdot)$. We then vary $c_R$ for $g_R(x)=c_R$ to obtain various level curves shown in (b), all of which approximate the corresponding density level sets well.
This is because the input preference pairs to rank-SVM incorporate density ordering information all over the support of $f$. Asymptotically $g_R$ preserves the ordering of density as will be shown in Sec.\ref{sec:analysis}.
This property of $g_R$ allows flexible false alarm control and does not need any re-training.


\subsubsection{Time Complexity}
For training, the rank computation step requires computing all pair-wise distances among nominal points $O(dn^2)$, followed by sorting for each point $O(n^2\log n)$.
So the training stage has the total time complexity $O(n^2(d+\log n)+T)$, where $T$ denotes the time of the pair-wise learning-to-rank algorithm.
At test stage, our algorithm only evaluates the SVM-type $g(\eta)$ on $\eta$ and does a binary search among $g(x_1),\ldots,g(x_n)$. The complexity is $O(s_R d+\log n)$, where $s_R$ is the number of support vectors, similar to $O(s_{oc}d)$ of one-class SVM, while nearest neighbor-based algorithms, K-LPE, aK-LPE or BP-KNNG \cite{ref:Manqi2009,ref:Jing2012,ref:knn_2011}, require $O(nd+n\log n)$ for testing one point. It is worth noting that $s_R\leq n$ comes from the ``support pairs'' within the input preference pair set, and is usually larger than $s_{OC}$.
Practically we observe that for most data sets $s_R$ is much smaller than $n$ in the experiment section, leading to significantly reduced test time compared to aK-LPE, as shown in Table.1.


\section{Analysis}\label{sec:analysis}
In this section we present some theoretical analysis of our ranking-based anomaly detection approach.
We first show that our approach is asymptotically consistent in that $r(\eta)$ converges to the $p$-value $p(\eta)$ as sample size approaches infinity.
We then provide a finite-sample generalization bound on the false alarm rate of our approach.

\subsection{Asymptotic Consistency}\label{subsec:asymptotic}
Our asymptotic analysis consists of three parts, respectively corresponding to the three main steps of our algorithm described in Sec.\ref{sec:main_algo}.

\noindent \textit{(1) Consistency of Rank Computation}

The rank of nominal samples based on average $K$-NN distance has been shown previously to converge to the $p$-value function \cite{ref:Jing2012}:
\begin{thm}\label{thm:consistency_step1}
Suppose the rank $r(x)$ is computed according to Eq.(\ref{eq:rank}) based on the average $K$-NN distance statistic Eq.(\ref{eq:aKnn}) among $\{x_1,\ldots,x_n\}$. With $K$ appropriately chosen, as $n\rightarrow\infty$,
\begin{equation}
    r(\eta) \rightarrow p(\eta).
\end{equation}
\end{thm}
This theorem establishes that asymptotically, the preference pairs generated as input to the learning-to-rank step are reliable, in the sense that any generated pair $(x,y)$ has the ``correct'' order as $p(x)>p(y)$, or equivalently $f(x)>f(y)$.

\noindent \textit{(2) Consistency of Rank SVM}

For simplicity we assume the preference pair set $\mathcal{P}$ contains all pairs over these $n$ samples.
Let $g_{\mathbf{x},\lambda}$  be the optimal solution to the Rank SVM Eq.(\ref{eq:ranksvm_standard}). If $L(x) = \max\{0,1-x\}$ denotes the
hinge loss, then this optimal solution satisfies

\begin{equation}\label{eq_emp_solution}
  g_{\bold{x},\lambda} = \arg \min_{g\in H} \lambda ||g||^2 + \ell_L(g;\bold{x}).
\end{equation}
where $\ell_L(g;\bold{x})$ is given by
\begin{equation}\label{eq_hinge_loss}
  \ell_L(g;\bold{x}) =\frac{1}{|\mathcal{P}|}\sum_{(i,j)\in \mathcal{P}} L( g(x_i)-g(x_j)).
\end{equation}

Let $\mathcal{H}_n$ denote a ball of radius $O(1/\sqrt{\lambda_n})$
in $H$. Let $C_k:= \sup_{x,t}|k(x,t)|$ with $k$ the rbf kernel associated
to $H$. Given $\epsilon>0$, we let
$N(\mathcal{H},\epsilon/4C_k)$ be the covering number
of $\mathcal{H}$ by disks of radius $\epsilon/4C_k$ (see appendix).
We first show that with appropriately chosen $\lambda$, as $n\rightarrow\infty$, $g$ is consistent in the following sense.

\begin{thm}\label{thm_convergence_surrogate}
Let $\lambda_n$ be appropriately chosen such that $\lambda_n\rightarrow 0$ and $\frac{\log N(\h_n,\epsilon/4C_k)}{n\lambda_n} \to 0$, as $n\rightarrow\infty$. Then we have
\begin{equation}
    E_{\bold{x}}[\ell_L(g_{\bold{x},\lambda};\bold{x})] \rightarrow  \inf_{g\in H} E_{\bold{x}}\left[ \ell_L(g;\bold{x}) \right].
\end{equation}
\end{thm}

We then establish that under mild conditions on the surrogate loss function, the solution minimizing the expected surrogate loss will asymptotically recover the correct preference relationships given by the density $f$.
\begin{thm}\label{thm_surrogate_condition}
Let $L$ be a non-negative, non-increasing convex surrogate loss function
that is differentiable at zero and satisfies $L'(0)<0$. If
\begin{equation*}
  g^* = \arg \min_{g\in H} \mathbb{E}_{\bold{x}} \left[ \ell(g;\bold{x}) \right],
\end{equation*}
then $g^*$ will correctly rank the samples according to their density, i.e.
$\forall x_i\neq x_j, r_i> r_j \implies g^*(x_i)>g^*(x_j)$, where $r_i=f(x_i)$, $r_j=f(x_j)$.
\end{thm}

The hinge-loss satisfies the conditions in the above theorem.
Combining Theorem \ref{thm_convergence_surrogate} and \ref{thm_surrogate_condition}, we establish that asymptotically, the rank-SVM step yields a ranker that preserves the preference relationship on nominal samples given by the nominal density $f$.

\noindent \textit{(3) Consistency of Test Stage Prediction} \\
\begin{cor}
Assume the non-flat condition of Lemma \ref{lem_p_mvset} holds. 
For a test point $\eta$, let $r(\eta)$ denote the rank computed by Eq.(\ref{eq:rank}) using the optimal solution of the rank-SVM step as $G$, as described in Algorithm 1.
Then for a given false alarm level $\alpha$, the decision region given in Algorithm 1 asymptotically converges to the ($1-\alpha$)-percentile minimum volume set decision region as in Eq.(\ref{eq:decision_mvset}).
\end{cor}

Thm.\ref{thm_surrogate_condition} and Lemma \ref{lem_order_consistency} yields the asymptotic consistency of $r(\cdot)$. Lemma \ref{lem_p_mvset} finishes the proof.

\subsection{Finite-Sample Generalization Result}\label{subsec:finite_sample}
Based on nominal samples $\{x_1,\ldots,x_n\}$, our approach learns a ranker $g_{n}$, and computes the values $g_{n}(x_1),\ldots,g_{n}(x_n)$. Let
$g_n^{(1)}\leq g_n^{(2)}\leq \cdots \leq g_n^{(n)}$ be the ordered permutation of these values.
For a test point $\eta$, we evaluate $g_{n}(\eta)$ and compute $r(\eta)$ according to Eq.(\ref{eq:rank}).
For a prescribed false alarm level $\alpha$, we define the decision region for claiming anomaly by
\begin{eqnarray*}
  R_\alpha &=& \{ x: \, r(x) \leq \alpha  \} \\
   &=&  \{ x: \, \sum_{j=1}^{n} \textbf{1}_{\{ g_n(x_j)\leq g_n(\eta) \}} \leq \alpha n \}  \\
   &=&  \{ x: \, g_n(\eta) < g_n^{ \lceil \alpha n \rceil } \}
\end{eqnarray*}
where $\lceil \alpha n \rceil$ denotes the ceiling integer of $\alpha n$.

We give a finite-sample bound on the probability that a newly drawn nominal point $x$ lies in $R_\alpha$. In the following Theorem, $\mathcal{F}$ denotes a
real-valued function class of kernel based linear functions (solutions
to an SVM-type problem) equipped with the $\ell_{\infty}$ norm over
a finite sample $\mathbf{x}=\{x_1,\dots,x_n\}$ :
\[
\lVert f \rVert _{\ell_{\infty}^{\mathbf{x}}} =\max_{x\in \mathbf{x}} |f(x)|.
\]
Moreover, $\mathcal{N}(\gamma,\mathcal{F},n)$ denotes a covering
number of $\mathcal{F}$ with respect to this norm (see \cite{ref:oc_svm2001}
for details).

\begin{thm}
Fix a distribution $P$ on $X$ and suppose $x_1,\dots,x_n$ are generated iid from $P$.
For $g\in\mathcal{F}$ let $g^{(1)}\leq g^{(2)}\leq \cdots \leq g^{(n)}$ be the ordered permutation of
$g(x_1),\dots,g(x_n)$. Then for such an $n$-sample, with probability $1-\delta$,
for any $g\in \mathcal{F}$, $1\leq m \leq n$ and sufficiently small $\gamma>0$,
\[
P\left\{x: g(x)< g^{(m)} - 2\gamma\right\} \leq \frac{m-1}{n} + \epsilon(n,k,\delta),
\]
where $\epsilon(n,k,\delta)=\frac{2}{n}(k+\log\frac{n}{\delta})$,
 $k=\lceil{\log\mathcal{N}(\gamma,\mathcal{F},2n)}\rceil$.
\end{thm}

\subsubsection*{Remark}
To interpret the theorem notice that the LHS is precisely the probability that a test point drawn from the nominal distribution has a score below the $\alpha\approx \frac{m-1}{n}$ percentile. We see that this probability is bounded from above by $\alpha$ plus an error term that asymptotically approaches zero. This theorem is true irrespective of $\alpha$ and so we have shown that we can simultaneously approximate multiple level sets. This theorem is similar to Theorem 12 of \cite{ref:oc_svm2001} in the second term of the upper bound. However, the generalization result for our approach applies to different quantiles $g^{(m)}$, or different values of $\alpha$, thus is a uniform upper bound on the empirical false alarm probability for different levels $\alpha$, while the result in \cite{ref:oc_svm2001} only applies for $g^{(1)}$, corresponding to one particular level set. This point is also illustrated in Fig.1.

\section{Experiments}
\label{sec:exp}
In this section, we carry out point-wise anomaly detection experiments on synthetic and real-world data sets. We compare our ranking-based approach against density-based methods BP-KNNG \cite{ref:knn_2011} and aK-LPE \cite{ref:Jing2012}, one-class SVM \cite{ref:oc_svm2001}, and another two state-of-art methods based on random sub-sampling, isolated forest \cite{ref:isolation_forest} (iForest) and massAD \cite{ref:massAD}.

\subsection{Implementation Details}
In our simulations, the Euclidean distance is used as distance metric for all candidate methods. For one-class SVM the lib-SVM codes \cite{ref:libsvm}  are used. The algorithm parameter and the RBF kernel parameter for one-class SVM are set using the same configuration as in \cite{ref:massAD}.
For iForest and massAD, we use the codes from the websites of the authors, with the same configuration as in \cite{ref:massAD}.
The $G(\cdot)$ statistic for our approach and aK-LPE is the average $k$-NN distance Eq.(\ref{eq:aKnn}) with fixed $k=10$. For BP-KNNG, the same $k$ is used and other parameters are set according to \cite{ref:knn_2011}.

For the rank-SVM step, we adapt the linear Ranking-SVM routine from \cite{ref:ranksvm_chapelle} to a kernelized version.
To generate preference pairs, we quantize the ranks of nominal points into $m$=3 levels $r(x_i)\rightarrow r_q(x_i)\in\{1,2,3\}$ and generate pairs $(i, j)\in\mathbb{P}$ whenever $r_q(x_i)>r_q(x_j)$.
We vary the rank-SVM parameter $C$ of Eq.(\ref{eq:ranksvm_standard}), $C\in \mathbb{C}=\{0.001,0.003,0.01,...,300,1000\}$, and the RBF kernel parameter $\sigma\in\Sigma=\{2^i\tilde{D}_K,\,i=-10,-9,\cdots,9,10\}$, where $\tilde{D}_K$ is the average $K$-NN distance over nominal samples.
We choose the parameter configuration through a 4-fold cross validation, and train a ranker $g(\cdot)$ with these parameters on the whole nominal set. Since anomalous data is unavailable at training time, we favor rankers that violate the preference relationships less on the nominal set. This $g(\cdot)$ is then adopted for test stage prediction.
All AUC performances are averaged over 5 runs.


\subsection{Synthetic Data sets}

We first apply our method to a Gaussian toy problem, where the nominal density is: \[ f_0 \sim 0.2 \mathcal{N} \left(
\left[ 5; 0 \right],
\left[ 1, 0 ; 0, 9 \right]
 \right)
+
0.8 \mathcal{N} \left(
\left[ -5 ; 0 \right],
\left[ 9 , 0; 0 , 1 \right]
 \right).
\]
The anomalous density $f_1$ is the uniform distribution within $\{(x,y):\,\,-18\leq x\leq 18, -18\leq y\leq 18\}$.

The empirical ROC curves of our method and one-class SVM along with the optimal Bayesian detector are shown in Fig.2. We can see from (a) that our approach performs fairly close to the optimal Bayesian classifier and much better than one-class SVM. Fig.2 (b) shows the level curves for the estimated ranks on the test data. The empirical level curves of rankAD approximate the level sets of the underlying nominal density quite well.

\begin{figure}[htbp]
\begin{centering}
\begin{minipage}[t]{.40\textwidth}
\includegraphics[width = 1\textwidth]{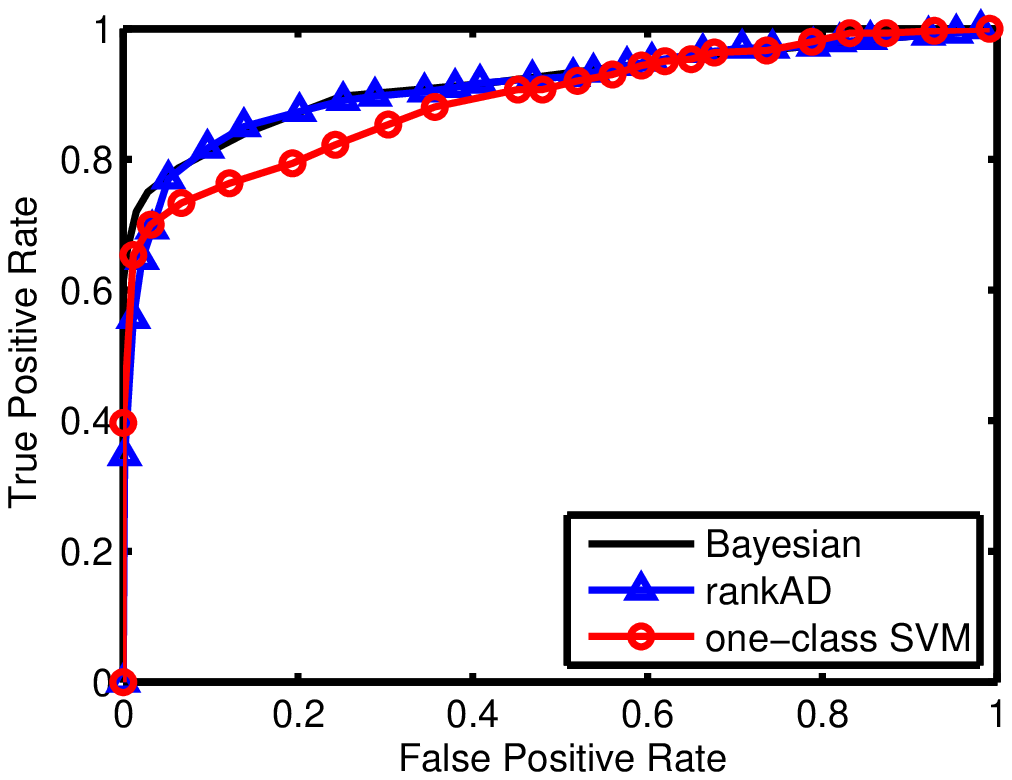}
\makebox[6.5 cm]{(a) ROC for 2-component Gaussian}\medskip
\end{minipage}
\begin{minipage}[t]{.40\textwidth}
\includegraphics[width = 1\textwidth]{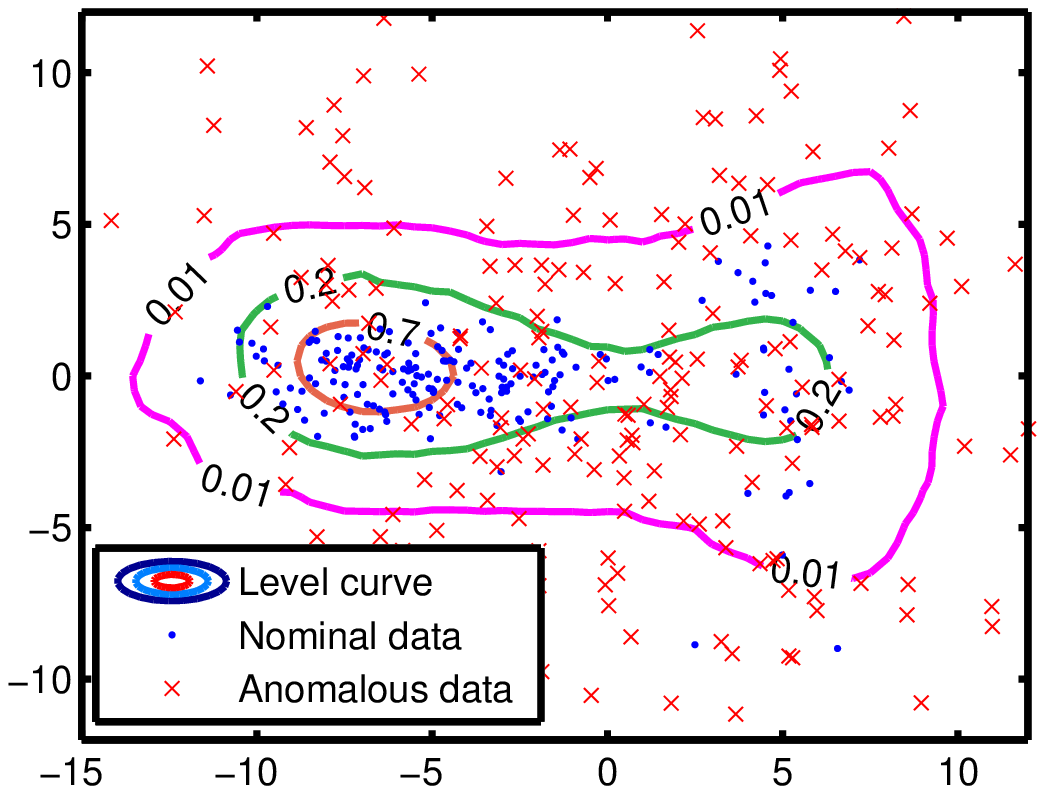}
\makebox[6.5 cm]{(b) Level curves of rankAD}\medskip
\end{minipage}
\caption{Performance on a synthetic data set: (a) ROC curve on a two-component Gaussian Mixture data. (b) Level sets for the estimated ranks. 600 training points are used for training. For test 500 nominal and 1000 anomalous points are used.}
\end{centering}
\label{fig:gaussian}
\end{figure}

\subsection{Real-world data sets}

We first illustrate ROC curves on two real-world data sets: the \textit{banknote} authentication data set and the \textit{magic} gamma telescope data set from the UCI repository \cite{ref:UCI}. We compare our approach to the two typical class of methods: density-based aK-LPE and one-class SVM. For the banknote data set, class with label 2 is regarded as nominal and other classes are anomalous. The testing time for 872 test points of aK-LPE, one-class SVM and our method are 0.078s, 0.02s and 0.031s (with 162/500 support vectors) respectively. As shown in Fig.3(a), our algorithm clearly outperforms one-class SVM. In fact, our method achieves 100\% true detection rate at a false positive rate of around 20\%, while one-class SVM achieves 100\% true detection rate at a false positive rate of 70\%.

\begin{figure}[htbp]
\begin{centering}
\begin{minipage}[t]{.40\textwidth}
\includegraphics[width = 1\textwidth]{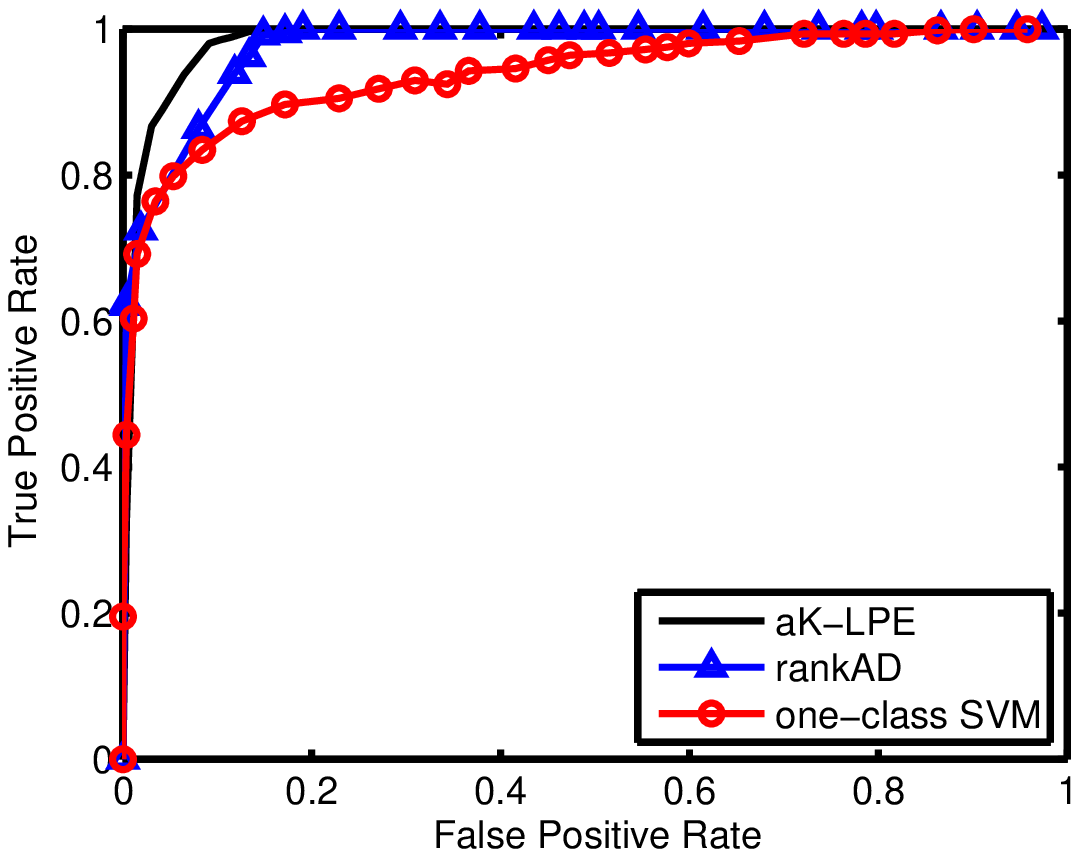}
\makebox[6.5 cm]{(a) Banknote}\medskip
\end{minipage}
\begin{minipage}[t]{.40\textwidth}
\includegraphics[width = 1\textwidth]{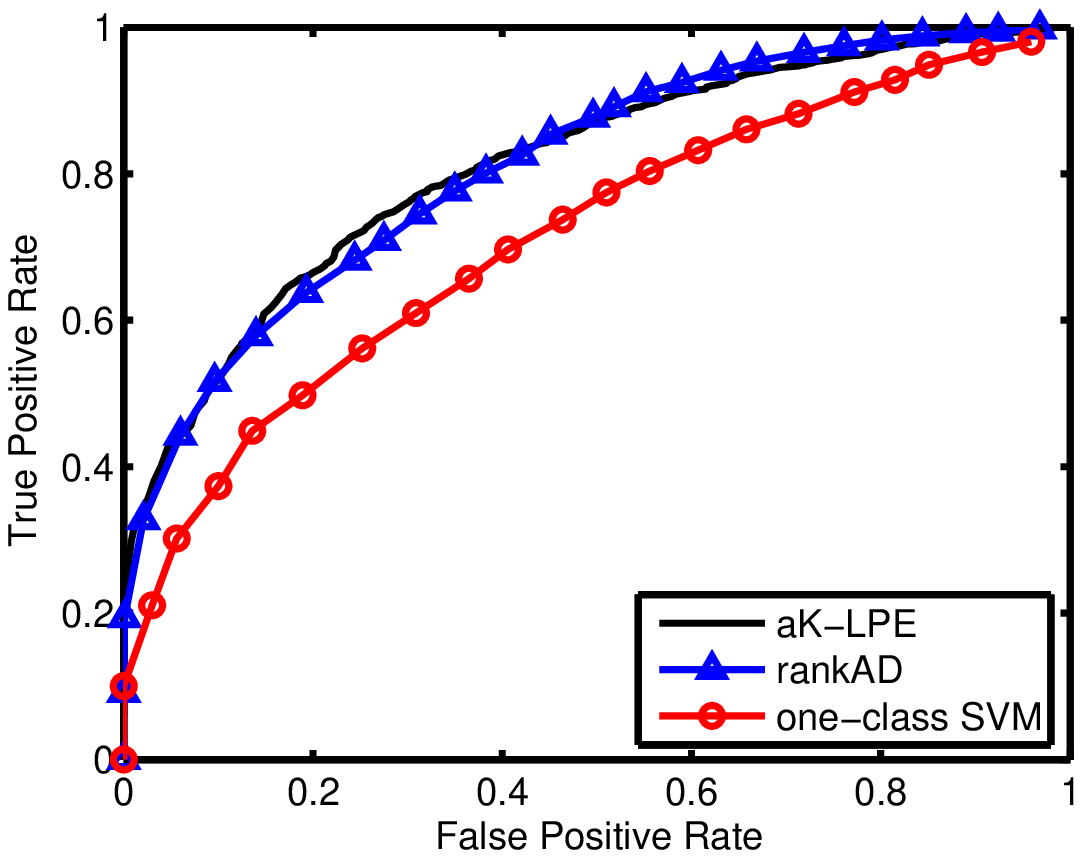}
\makebox[6.5 cm]{(b) Magic Gamma}\medskip
\end{minipage}
\caption{The ROC curves for aK-LPE, one-class SVM and the proposed method on different data sets. (a) Banknote authentication, class ``2'' (nominal) vs. others, 5-dim, 500 training points, 872 test points (262 nominal). (b) Magic Gamma Telescope, gamma particles (nominal) vs. background, 10-dim, 1500 training points, 4000 test points (1000 nominal).}
\end{centering}
\label{fig:ROCs}
\end{figure}

The Magic gamma telescope data set is an image data set used to classify high energy gamma particles from cosmic rays in an atmospheric telescope. 10 attributes of the observed images are used as input features. Here we regard all gamma particles as nominal data and background cosmic rays as anomaly. The testing time for 4000 test points of aK-LPE, one-class SVM and our method are 0.42s, 0.01s and 0.01s (with 41/1500 support vectors) respectively. Fig.3(b) demonstrates that our method significantly outperforms one-class SVM and is comparable to aK-LPE but with significantly smaller test time.

\begin{table}[!htbp]
\caption{ Data characteristics of the data sets used in experiments. $N$ is the total number of instances. $d$ the dimension of data. The percentage in brackets indicates the percentage of anomalies among total instances. }
\begin{center}
\begin{tabular}{|c||c|c|c|}
  \hline
  data sets     & $N$  &  $d$  & anomaly class  \\
  \hline\hline
  Annthyroid     & 6832  &  6  &  classes 1,2(7\%) \\    \hline
  Forest Cover   & 286048  &  10  &  class 4(0.9\%) vs. class 2  \\ \hline
  HTTP           & 567497  &  3   &  attack(0.4\%)  \\ \hline
  Mamography     & 11183   &  6   &  class 1(2\%)  \\ \hline
  Mulcross       & 262144  &  4   &  2 clusters(10\%)  \\ \hline
  Satellite      & 6435    &  36  &  3 smallest classes(32\%)  \\ \hline
  Shuttle        & 49097   &  9   &  classes 2,3,5,6,7(7\%)  \\ \hline
  SMTP           & 95156  &  3   &  attack (0.03\%)   \\
  \hline
\end{tabular}
\end{center}
\label{tab:data_sets}
\end{table}

We then conduct experiments on several other real data sets used in \cite{ref:isolation_forest} and \cite{ref:massAD}, including 2 network intrusion data sets HTTP and SMTP from \cite{ref:Yamanishi00}, Annthyroid, Forest Cover Type, Satellite, Shuttle from UCI repository \cite{ref:UCI}, Mammography and Mulcross from \cite{ref:Rocke96}. Table \ref{tab:data_sets} illustrates the characteristics of these data sets.

\begin{table*}[!tp]
\caption{Anomaly detection AUC performance and test stage time of various methods.}
\begin{center}
\begin{tabular}{|c|c||c|c|c|c|c|c|}
  \hline
  \multicolumn{2}{|c||}{Data Sets} &   rankAD  &   oc-svm  &   BP-KNNG   & aK-LPE & iForest & massAD \\
  \hline\hline
  \multirow{8}{*}{AUC}
  &    Annthyroid        & 0.844         & 0.681   & 0.823   &  0.753  & {\bf 0.856}   & 0.789 \\
  &   Forest Cover      & {\bf 0.932}   & 0.869   & 0.859   &  0.876  & 0.853         & 0.895 \\
  &    HTTP              & {\bf 0.999}   & 0.998   & 0.995   & {\bf 0.999}  & 0.986         & 0.995 \\
  &    Mamography        & {\bf 0.909}   & 0.863   & 0.868   &  0.879  & 0.891         & 0.701 \\
  &    Mulcross          & {\bf 0.998}   & 0.970   & 0.994   & {\bf 0.998}  & 0.971         & 0.998 \\
  &    Satellite         & {\bf 0.885}   & 0.774   & 0.872   &  0.884  & 0.812         & 0.692 \\
  &    Shuttle           & {\bf 0.996}   & 0.975   & 0.985   &  0.995  & 0.992         & 0.992 \\
  &   SMTP              & {\bf 0.934}   & 0.751   & 0.892   &  0.900  & 0.869         & 0.859 \\
  \hline
  \multirow{8}{*}{test time}
    & Annthyroid        & 0.338  & 0.281 & 2.171	& 0.917  & 1.384  & 0.030 \\
    & Forest Cover      & 1.748  & 1.638 & 2.185	& 13.41  & 7.239  & 0.483 \\
    & HTTP              & 0.187  & 0.376 & 2.391	& 11.04  & 5.657  & 0.384 \\
    & Mamography        & 0.237  & 0.223 & 0.281	& 1.443  & 1.721  & 0.044 \\
    & Mulcross          & 2.732  & 2.272 & 3.772 	& 13.75  & 7.864  & 0.559 \\
    & Satellite         & 0.393  & 0.355 & 0.776 	& 1.199  & 1.435  & 0.030 \\
    & Shuttle           & 1.317  & 1.318 & 2.404 	& 7.169  & 4.301  & 0.186 \\
    & SMTP              & 1.116  & 1.105 & 1.912 	& 11.76  & 5.924  & 0.411 \\
  \hline
\end{tabular}
\end{center}
\label{tab:real_AUC}
\end{table*}

We randomly sample 2000 nominal points for training. The rest of the nominal data and all of the anomalous data are held for testing. Due to memory constraint, at most 80000 nominal points are used at test time. The time for testing all test points and the AUC performance are reported in Table \ref{tab:real_AUC}.

We observe that while being faster than BP-KNNG, aK-LPE and iForest, and comparable to oc-SVM during test stage, our approach also achieves very good performance for all data sets.
The density based aK-LPE has somewhat good performance, but its test-time degrades significantly with training set size. The other density based BP-KNNG has less test time compared to aK-LPE since it uses a subset of the training samples, however its performance is not comparable to rankAD. massAD is fast at test stage, but has poor performance for several data sets.
Overall, our approach is competitive in both AUC performance and test time compared to other state-of-art algorithms.

\section{Conclusions}
\label{sec:con}
In this paper, we propose a novel anomaly detection framework based on RankSVM. We combine statistical density information with a discriminative ranking procedure.
Our scheme learns a ranker over all nominal samples based on the $k$-NN distances within the graph constructed from these nominal points. This is achieved through a pair-wise learning-to-rank step, where the inputs are preference pairs $(x_i,x_j)$. The preference relationship for $(x_i,x_j)$ takes a value one if the nearest neighbor based score for $x_i$ is larger than that for $x_j$. Asymptotically this preference models the situation that data point $x_i$ is located in a higher density region relative to $x_j$ under nominal distribution.
We then show the asymptotic consistency of our approach, which allows for flexible false alarm control during test stage.
We also provide a finite-sample generalization bound on the empirical false alarm rate of our approach.
Experiments on synthetic and real data sets demonstrate our approach has superior performance as well as low test time complexity.


%

\appendix[Supplementary: Proofs of Theorems]

We fix an RKHS $H$ on the input space $X\subset\mathbb{R}^d$
with an RBF kernel $k$.
Let $\mathbf{x}=\{x_1,\dots,x_n\}$ be a set of objects to be ranked
in $\mathbb{R}^d$ with labels $\mathbf{r}= \{r_1,\dots,r_n\}$.
Here $r_i$ denotes the label of $x_i$, and $r_i\in \mathbb{R}$.
We assume $\mathbf{x}$ to be a random variable distributed according
to $P$, and $\mathbf{r}$ deterministic.   Throughout
$L$ denotes the hinge loss.

The following notation will be useful in the proof of Theorem 4.
Define the $L$-$risk$ of $f\in H$ as
\[
\RP (f) = E_{\bold{x}}\ell_L(f;\bold{x})
\]
where
\[
\ell_L(f;\bold{x})=\sum_{i,j:r_i>r_j}D(r_i,r_j)L(f(x_i)-f(x_j))
\]
and $D(r_i,r_j)$ is some positive weight function such as
$1/|\mathcal{P}|$.
The smallest possible $L$-risk in $H$ is denoted $\RP$.
The {\it regularized} $L$-$risk$ is
\begin{equation}\label{regularize}
\Rp^{\text{reg}}(f):=\lambda \lVert f\rVert^2+\mathcal{R}_{L,P}(f),
\end{equation}
 $\lambda >0$. If $P$ is
the empirical measure with respect to $T\in (X\times X)^{\e}$,
$\e=\binom{n}{2}$,
we write $\RT(f)$ and $\Rt^{\text{reg}}(f)$ for the associated risks:
\[
\RT(f)= \sum_{i.j: r_i>r_j}D(r_i,r_j)L \left( f(x_i)-f(x_j) \right)
\]
\[
\Rt^{\text{reg}}(f)= \lambda ||f||^2 + \RT(f)
\]

\subsection{Proof of Theorem 4}

\begin{Proof}
Let us outline the argument. In \cite{ref:Steinwart2001}, the author shows that there exists a $f_{P,\lambda}\in H$
minimizing (\ref{regularize}):
\begin{lemma}\label{convex}
For all Borel probability measures $P$ on $X\times X$ and
all $\lambda >0$, there is an $f_{P,\lambda} \in H$ with
\[
\Rp^{\text{reg}}(f_{P,\lambda}) = \inf_{f\in H} \Rp^{\text{reg}}(f)
\]
such that $\lVert f_{P,\lambda} \rVert =O(1/\sqrt{\lambda})$.
\end{lemma}

Next, a simple argument shows that
\[
\lim_{\lambda\to 0} \Rp^{\text{reg}}(f_{P,\lambda})= \RP.
\]

Finally, we will need a concentration inequality to relate the $L$-risk
of $f_{P,\lambda}$ with the empirical $L$-risk of $f_{T,\lambda}$.
We then derive consistency using the following argument:
\begin{align*}
\RP(f_{T,\lambda_n})
& \leq  \lambda_n \lVert f_{T,\lambda_n}\rVert^2+\mathcal{R}_{L,P}(f_{T,\lambda_n})\\
& \leq \lambda_n \lVert f_{T,\lambda_n}\rVert^2+\mathcal{R}_{L,T}(f_{T,\lambda_n})+\delta/3  \\
& \leq  \lambda_n \lVert f_{P,\lambda_n}\rVert^2+\mathcal{R}_{L,T}(f_{P,\lambda_n})+\delta/3 \\
& \leq  \lambda_n \lVert f_{P,\lambda_n}\rVert^2+\mathcal{R}_{L,P}(f_{P,\lambda_n})+2\delta/3  \\
& \leq  \mathcal{R}_{L,P}+\delta
\end{align*}

where $\lambda_n$ is an appropriately chosen sequence $\to 0$,
and $n$ is large enough. The second and fourth inequality holds due to Concentration Inequalities, and the last one holds since $\lim_{\lambda \to 0} \Rp^{\text{reg}}(f_{P,\lambda})=\mathcal{R}_{L,P}$.

We now prove the appropriate concentration inequality \cite{ref:Smale2001}.
Recall $H$ is an RKHS with
smooth kernel $k$; thus the inclusion $I_{k}: H\to C(X)$ is compact, where
$C(X)$ is given the $\lVert \cdot \rVert_{\infty}$-topology. That is, the
``hypothesis space'' $\mathcal{H}:= \overline{I_k(B_R)}$ is compact in $C(X)$,
where $B_R$ denotes the ball of radius $R$ in $H$. We denote by $N(\mathcal{H},
\epsilon)$ the covering number of $\mathcal{H}$ with disks of radius $\epsilon$.
We prove the following inequality:

\begin{lemma}
For any probability distribution $P$ on $X\times X$,
\begin{equation*}
P^{\e}\{T\in (X\times X)^{\e}:\sup_{f\in \h} | \RT(f)-\RP(f)| \geq \epsilon \}\leq
2N(\h,\epsilon/4C_k)\exp\left(
\frac{-\epsilon^2n}{2(1+2\sqrt{C_k}R)^2}\right),
\end{equation*}

where $C_k := \sup_{x,t}|k(x,t)|$.
\end{lemma}

\begin{Proof}
Since $\h$ is compact, it has a finite covering number. Now suppose $\h=
D_1\cup \cdots \cup D_{\ell}$ is any finite covering of $\h$. Then
\begin{equation*}
\text{Prob}\{\sup_{f\in \h} | \RT(f)-\RP(f)| \geq \epsilon \} \leq  \sum_{j=1}^{\ell}
\text{Prob}\{\sup_{f\in D_j} | \RT(f)-\RP(f)| \geq \epsilon \}
\end{equation*}
so we restrict attention to a disk $D$ in $\h$ of appropriate radius $\epsilon$.

Suppose $\lVert f-g \rVert_{\infty}\leq \epsilon$. We want to show that the
difference
\[
|(\RT(f)-\RP(f))-(\RT(g)-\RP(g))|
\]
is also small. Rewrite this quantity as
\[
 |(\RT(f)-\RT(g))-E_{\bold{x}}[\RT(g)-\RT(f)]|.
 \]
 Since $\lVert f-g \rVert_{\infty}\leq \epsilon$, for $\epsilon$ small enough we have
\begin{align*}
\max\{0,1-(f(x_i)-f(x_j))\}-\max\{0,1-(g(x_i)-g(x_j))\} & = \max\{0,(g(x_i)-g(x_j)-f(x_i)+f(x_j))\} \\
& = \text{max}\{0,\langle g-f, \phi(x_i)-\phi(x_j)\rangle\}.
\end{align*}
Here $\phi:X\to H$ is the feature map, $\phi(x):=k(x,\cdot)$.
 Combining this with the Cauchy-Schwarz inequality, we have
\begin{eqnarray*}
 |(\RT(f)-\RT(g))-E_{\bold{x}}[\RT(g)-\RT(f)]| & \leq  \frac{2}{n^2}(2n^2\lVert f-g \rVert_{\infty}C_k) & \leq 4C_k\epsilon,
\end{eqnarray*}
where $C_k:= \sup_{x,t} |k(x,t)|$. From this inequality it follows that
\begin{equation*}
|\RT(f)-\RP(f)| \geq (4C_k+1)\epsilon  \implies |(\RT(g)-\RP(g))| \geq \epsilon.
\end{equation*}
We thus choose to cover $\h$ with disks of radius $\epsilon/4C_k$, centered at
$f_1,\dots,f_{\ell}$. Here $\ell= N(\h,\epsilon/4C_k)$ is the covering number
for this particular radius. We then have
\begin{equation*}
\sup_{f\in D_j}|(\RT(f)-\RP(f))|\geq 2\epsilon  \implies |(\RT(f_j)-\RP(f_j))|\geq \epsilon.
\end{equation*}
Therefore,
\begin{equation*}
 \text{Prob}\{\sup_{f\in \h} | \RT(f)-\RP(f)| \geq 2\epsilon \}  \leq
 \sum_{j=1}^n \text{Prob}\{ | \RT(f_j)-\RP(f_j)| \geq \epsilon \}
 \end{equation*}
The probabilities on the RHS can be bounded using McDiarmid's inequality.

Define the random variable $g(x_1,\dots,x_n) :=\mathcal{R}_{L,T}(f)$, for fixed $f\in H$.
We need to verify that $g$ has bounded differences. If we change one of the
variables, $x_i$, in $g$ to $x_i'$, then at most $n$ summands will change:
\begin{align*}
|g(x_1,\dots,x_i,\dots,x_n)-g(x_1,\dots,x_i',\dots,x_n)|
& \leq \frac{1}{n^2}2n\sup _{x,y} |1-(f(x)-f(y))| \\
& \leq \frac{2}{n}+\frac{2}{n}\sup_{x,y}|f(x)-f(y)|\\
& \leq \frac{2}{n}+\frac{4}{n}\sqrt{C_k}\lVert f \rVert.
\end{align*}
Using that $\sup_{f\in \mathcal{H}}\lVert f \rVert\leq R$,
McDiarmid's inequality thus gives
\begin{equation*}
\text{Prob}\{\sup_{f\in \h} | \RT(f)-\RP(f)| \geq \epsilon \}
 \leq 2N(\h,\epsilon/4C_k)\exp\left(
\frac{-\epsilon^2n}{2(1+2\sqrt{C_k}R)^2}\right).
\end{equation*}
\end{Proof}

We are now ready to prove Theorem 4. Take $R=\lVert f_{P,\lambda} \rVert$ and apply this result to $f_{P,\lambda}$:
\begin{equation*}
\text{Prob}\{| \RT(f_{P,\lambda})-\RP(f_{P,\lambda})| \geq \epsilon \}  \leq
2N(\h,\epsilon/4C_k)\exp\left(
\frac{-\epsilon^2n}{2(1+2\sqrt{C_k}\lVert f_{P,\lambda} \rVert)^2}\right).
\end{equation*}
Since $\lVert f_{P,\lambda_n} \rVert =O(1/\sqrt{\lambda_n})$, the RHS converges to 0
so long as $\dfrac{n\lambda_n}{\log N(\h,\epsilon/4C_k)} \to \infty$ as $n\to \infty$.
This completes the proof of Theorem 4.
\end{Proof}

\subsection{Proof of Theorem 5}
\begin{Proof} Our proof follows similar lines of Theorem 4 in \cite{ref:RDPS2012}.
Assume that $g(x_i) < g(x_j)$, and define a function $g'$ such that
$g'(x_i)=g(x_j)$, $g'(x_j)=g(x_i)$, and $g'(x_k)=g(x_k)$ for all $k\neq i,j$.
We have $\RP(g')-\RP(g)=E_{\bold{x}}(A(\bold{x}))$, where

\begin{eqnarray*}
 A(\bold{x})
   = \sum_{k : r_j<r_i<r_k}[D(r_k,r_j)-D(r_k,r_i)]  [L(g(x_k)-g(x_i))-L(g(x_k)-g(x_j))]
  \\ +  \sum_{k : r_j<r_k<r_i}D(r_i,r_k)[L(g(x_j)-g(x_k))-L(g(x_i)-g(x_k))]
  \\ +  \sum_{k : r_j<r_k<r_i}D(r_k,r_j)[L(g(x_k)-g(x_i))-L(g(x_k)-g(x_j))]
  \\ +  \sum_{k : r_j<r_i<r_k}[D(r_k,r_j)-D(r_k,r_i)][L(g(x_k)-g(x_i))-L(g(x_k)-g(x_j))]
  \\ +  \sum_{k : r_j<r_i<r_k}[D(r_i,r_k)-D(r_j,r_k)][L(g(x_j)-g(x_k))-L(g(x_i)-g(x_k))]
  \\ +  (L(g(x_j)-g(x_i))-L(g(x_i)-g(x_j)))D(r_i,r_j).
  \end{eqnarray*}
  Using the requirements of the weight function $D$ and the assumption that $L$
  is non-increasing and non-negative, we see that all six sums in the above
  equation for $A(\bold{x})$ are negative. Thus $A(\bold{x})<0$, so
  $\RP(g')-\RP(g)=E_{\bold{x}}(A(\bold{x}))<0$, contradicting the minimality
  of $g$. Therefore $g(x_i)\geq g(x_j)$.

  Now we assume that $g(x_i)=g(x_j)=g_0$. Since $\RP(g)=\inf_{h\in H}\RP(h)$,
  we have $\left. \dfrac{\partial{\ell_L(g;x)}}{\partial{g(x_i)}}\right|_{g_0}=A=0,$ and
  $\left. \dfrac{\partial{\ell_L(g;x)}}{\partial{g(x_j)}}\right|_{g_0}=B=0$, where
  \begin{eqnarray*}
  A=\sum_{k : r_j < r_i < r_k} D(r_k, r_i) [ -L'(g(x_k)-g_0)]+
    \sum_{k : r_j < r_k< r_i} D(r_i, r_k) L'(g_0-g(x_k)) +\\
  \sum_{k : r_k < r_j < r_i} D(r_i, r_k)  L'(g_0-g(x_k))+D(r_i,r_j)[-L'(0)].
  \end{eqnarray*}
  \begin{eqnarray*}
  B=\sum_{k : r_j < r_i < r_k} D(r_k, r_j) [ -L'(g(x_k)-g_0)]+
  \sum_{k : r_j < r_k< r_i} D(r_k, r_j) L'(g_0-g(x_k)) +\\
  \sum_{k : r_k < r_j < r_i} D(r_j, r_k)  L'(g_0-g(x_k))+D(r_i,r_j)[-L'(0)].
  \end{eqnarray*}
  However, using $L'(0)<0$ and the requirements of $D$ we have
  \[
  A-B\leq 2L'(0)D(r_i,r_j)<0,
  \]
  contradicting $A=B=0$.
\end{Proof}

\subsection{Proof of Theorem 6}

To prove Theorem 6 we need the following lemma \cite{ref:vapnik1979}:

\begin{lemma}\label{vapnik}
Let $\X$ be a set and $S$ a system of sets in $\X$, and $P$ a probability
measure on $S$. For $\x\in \X^{n}$ and $A\in S$, define $\nu_{\x}(A):= |\x\cap A|/n$.
If $n>2/\epsilon$, then
\begin{equation*}
P^{n}\left\{ \x : \sup_{A\in S} |\nu_{\x}(A)- P(A)|>\epsilon\right\}  \leq
2P^{2n}\left\{ \x\x' : \sup_{A\in S} |\nu_{\x}(A)-\nu_{\x'}(A)|>\epsilon/2\right\}.
\end{equation*}
\end{lemma}

\begin{Proof}
Consider the event
\begin{equation*}
J:=  \Biggl\{  \x \in \X^n : \exists f\in \F, P\{ x : f(x) < f^{(m)} - 2\gamma \}  > \frac{m-1}{n}+\epsilon \Biggr\}.
\end{equation*}

We must show that $P^n(J)\leq \delta$ for $\epsilon = \epsilon(n,k,\delta)$.
Fix $k$ and apply lemma \ref{vapnik} with
\[
A=\{ x : f(x) < f^{(m)} - 2\gamma \}
\]
with $\gamma$ small enough so that
\[
\nu_{\x}(A)= |\{x_j\in \x : f(x_j) < f^{(m)} - 2\gamma\}|/n=
 \frac{m-1}{n}.
 \]
We obtain
\[
P^n(J)\leq 2P^{2n} \Biggl\{ \x\x' :  \exists f\in \F, | \{ x_j' \in \x' : f(x_j')  < f^{(m)} - 2\gamma\}| > \epsilon n/2 \Biggr\}.
\]
The remaining portion of the proof follows as Theorem 12 in \cite{ref:oc_svm2001}.
\end{Proof}

%
%

\ifCLASSOPTIONcaptionsoff
  \newpage
\fi



%
%
%

\bibliographystyle{IEEEtran}
\bibliography{IEEEabrv,ranksvm}

%

%
%
%
%
%
%
%
%




\end{document}